\documentclass{article}
\usepackage{spconf,amsmath,graphicx}
\usepackage[linesnumbered,boxed]{algorithm2e}
\usepackage{multirow}
\usepackage{booktabs}
\usepackage{subfigure}
\usepackage{xcolor}
\usepackage[colorlinks,linkcolor=blue, citecolor=black]{hyperref}

\title{Y-Net: MULTI-SCALE FEATURE AGGREGATION NETWORK WITH WAVELET STRUCTURE SIMILARITY LOSS FUNCTION FOR SINGLE IMAGE DEHAZING}
\name{Hao-Hsiang Yang$^{1}$ \qquad Chao-Han Huck Yang$^{2}$ \qquad Yi-Chang James Tsai$^{2}$}
\address{$^{1}$ASUS Intelligent Cloud Services, Taiwan\\  $^{2}$School of Electrical and Computer Engineering, Georgia Institute of Technology, Atlanta, GA, USA }

\begin{document}
\maketitle
\begin{abstract}
Single image dehazing is the ill-posed two-dimensional signal reconstruction problem. Recently, deep convolutional neural networks (CNN) have been successfully used in many computer vision problems. In this paper, we propose a Y-net that is named for its structure. This network reconstructs clear images by aggregating multi-scale features maps. Additionally, we propose a Wavelet Structure SIMilarity (W-SSIM) loss function in the training step. In the proposed loss function, discrete wavelet transforms are applied repeatedly to divide the image into differently sized patches with different frequencies and scales. The proposed loss function is the accumulation of SSIM loss of various patches with respective ratios. Extensive experimental results demonstrate that the proposed Y-net with the W-SSIM loss function restores high-quality clear images and outperforms state-of-the-art algorithms. Code and models are available at \href{https://github.com/dectrfov/Y-net}{https://github.com/dectrfov/Y-net}

\end{abstract}
\begin{keywords}
Single image dehazing, Y-net, discrete wavelet transform, structure similarity, multi-scale feature aggregation
\end{keywords}
\section{Introduction}
\label{sec:intro}
Single image dehazing is the two-dimensional signal reconstruction problem and intends to restore the unknown clear image given a hazy or foggy image. The image dehazing model \cite{mccartney1976optics} is formulated as follows:
\begin{equation}
I(x)=J(x)t(x)+A(1-t(x))
\label{eq:1}
\end{equation}
where \textit{$\ I(x) $} is the observed image, \textit{$\ J(x) $} is the clear image we need to estimate, \textit{$A$} is the global atmospheric light, and \textit{$t(x)$} is the medium transmission. Assuming that the atmosphere is homogeneous, we express the transmission as {$t(x) = e^{-\beta d(x) }$}, where {$\beta$} is the scattering coefficient of atmosphere, and {$d(x)$} is the scene depth. Since all variables in Equation (1) are unknown except the hazy image \textit{$\ I(x)$}, image dehazing is an ill-posed problem.

Previously, many efforts on developing visual priors capture deterministic and statistical properties of hazy images.
In \cite{zhu2015fast}, color attenuation prior is proposed to recover the depth information. Berman \emph{et al.} \cite{berman2016non} find colors in haze-free images can be well approximated by a few hundred distinct colors that form tight clusters in RGB space. Garldran \emph{et al.} \cite{galdran2018duality} prove that Retinex on inverted intensities is a solution to the image dehazing problem. Cho \emph{et al.} \cite{cho2018model} propose the multi-band fusion approach using several inputs derived from the original image, and present balanced image enhancement while elaborating image details.

Deep learning frameworks also show competitive results in the dehazing problems. In \cite{ren2016single}, the multi-scale deep neural network is proposed to learn a non-linear mapping. Li \emph{et al.} \cite{li2017aod} reformulate the atmospheric scattering model and present a lightweight model to produce clear images from hazy images. Recently, several deep-learning-based U-Nets methods~\cite{ronneberger2015u, yang2018auto, yang2018novel} have also been proposed. In \cite{yang2019wavelet}, authors replace down-sampling with the discrete wavelet transform in U-net~\cite{ronneberger2015u} to estimate dehazed images.

Similarly, we consider neural networks for image dehazing, since the dehazing model is only a crude approximation and CNN can capture more detailed features from hazy images.
We think reconstructed images are composed of different scale feature maps and propose the Y-net. The Y-net aggregates various size feature maps that are up-sampled to identical size and convolves them by 1{$\times$}1 kernels to restore clear images. Thus, all feature maps are comprised of the output images, and weights of all layers are updated effectively. 
In this paper, we show that applying existing encoder-decoder structures like U-net \cite{ronneberger2015u} or cascaded refinement networks \cite{nah2017deep} cannot produce optimal results. Our Y-net, on the contrary, amplifies the merit of various CNN structures and yields the feasibility for single image dehazing.

Not only network structures but loss functions impact accuracy. In \cite{zhao2016loss}, the loss function based on the Structure SIMilarity (SSIM) index \cite{wang2004image} is proposed for image restoration. The authors demonstrate that when the network architecture is unchanged, the quality of the results improves significantly with better loss functions. Therefore, we extend the SSIM loss by combining it with the discrete wavelet transform (DWT) and propose the wavelet SSIM (W-SSIM) loss. Images are divided into many patches by DWT with various frequencies. Then, the SSIM loss of each patch is calculated, and the weights of each loss are adjustable to preserve high-frequency details and prevent halo artifacts. Experimental results show that the proposed loss function achieves improved reconstructions with an identical network.

The contributions of this paper are two-fold. First, we propose a deep end-to-end trainable Y-net to reconstruct clear images without estimating any priors on atmospheric light and scene transmission. We demonstrate the utility and effectiveness of the Y-net for single image dehazing on the synthetic hazy image dataset. Second, the W-SSIM loss is proposed to improve accuracy. We show that the proposed dehazing model performs favorably against the related methods.
\section{METHODOLOGY}
\label{sec:format}
\subsection{Y-net:~Multi-scale feature aggregation network}
The proposed Y-net architecture mainly consists
of three modules: down-sampling, up-sampling, and aggregation parts. The first two modules are similar to the classical U-net \cite{ronneberger2015u}. Our Y-net uses convolutions and down-sampling to extract multi-scale features. After bottom features are computed, convolutions and up-sampling are adopted to magnify the size of feature maps. The original U-net adopts the final up-sampled feature map to represent reconstructed images. Different from the U-net, we think reconstructed images are composed of different scale feature maps, and propose the aggregation module to combine each feature map to reconstruct clear images. Each feature map is adjusted to identical size by up-sampling and convolved by 1{$\times$}1 kernels to restore clear images. The 1{$\times$}1 convolution is seen as the operation of the weighted sum of various feature maps. Thus, all feature maps are comprised of the output images, and weights of all layers are updated effectively, which also avoids vanishing gradient.
Our input is the 3-channel image, convolutions, whose size are 3{$\times$}3 with a stride of two are adopted to down-sampling, the whole network architecture is shown in Fig. 1.
\begin{figure}[ht!]
\label{fig:1}
\begin{minipage}[b]{1.0\linewidth}
  \centering
  \centerline{\includegraphics[width=0.9\linewidth]{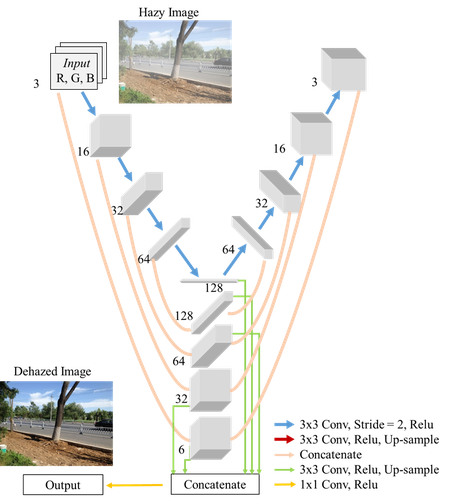}}
    \caption{{The overview of our proposed Y-net. The clear image is composed of multi-scale feature maps from the hazy image. The digits under the blocks mean the numbers of channels.}}
    \vspace{-0.5cm}
\end{minipage}
\end{figure}
\subsection{Wavelet SSIM loss}
In the two dimensional discrete wavelet transform (DWT), for instance, four filters, {$f_{LL}$}, {$f_{HL}$}, {$f_{LH}$} and {$f_{HH}$}, are used to convolve with an image. The convolution results are then down-sampled. The 2-D DWT has separable property, so four filters are obtained by multiplication of the scaling function and the wavelet function. Scaling function is seen as the low-pass filter, and the wavelet function is seen as the high-pass filter. The block diagram of the DWT and the results are showing in Fig. 2, and the formula is represented as follows:

\begin{figure}[!htb]
\centering
\subfigure[]{
\begin{minipage}[t]{0.5\linewidth}
\centering
\includegraphics[width=1.5in]{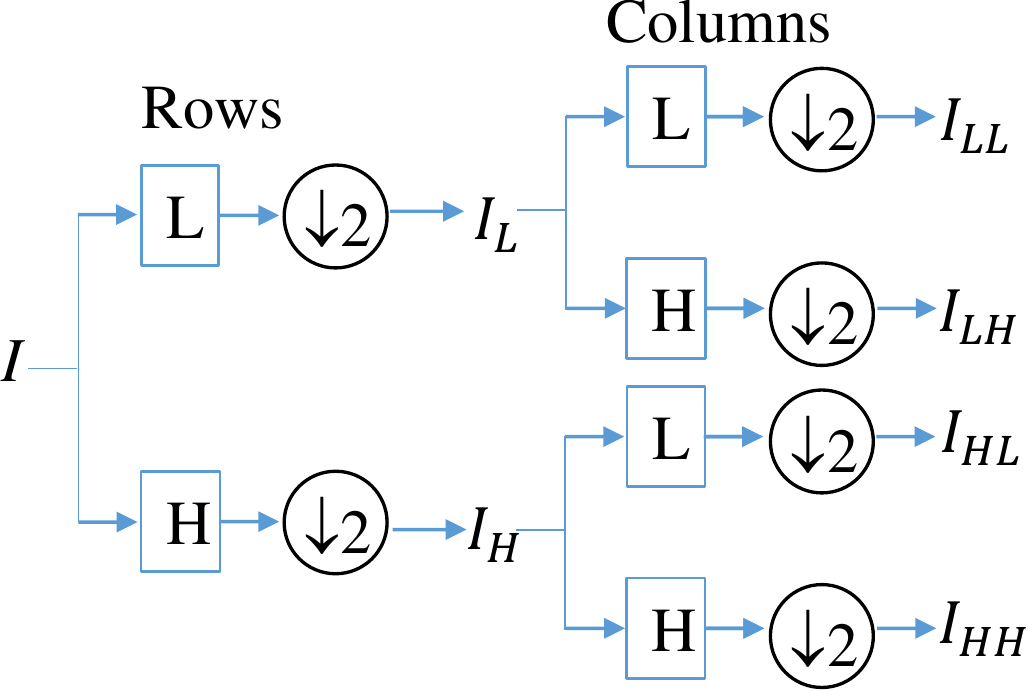}
\end{minipage}%
}%
\subfigure[]{
\begin{minipage}[t]{0.5\linewidth}
\centering
\includegraphics[width=1.5in]{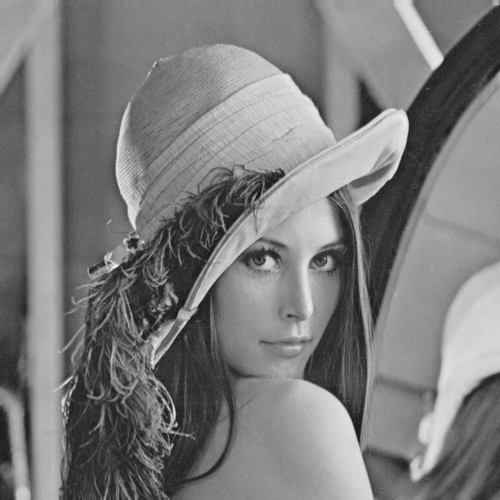}
\end{minipage}%
}%

\subfigure[]{
\begin{minipage}[t]{0.5\linewidth}
\centering
\includegraphics[width=1.5in]{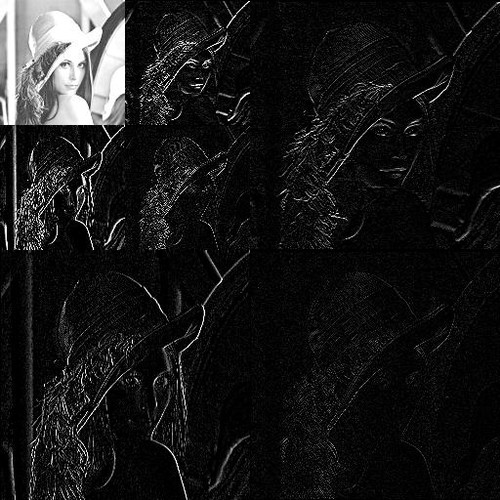}
\end{minipage}%
}%
\subfigure[]{
\begin{minipage}[t]{0.5\linewidth}
\centering
\includegraphics[width=1.5in]{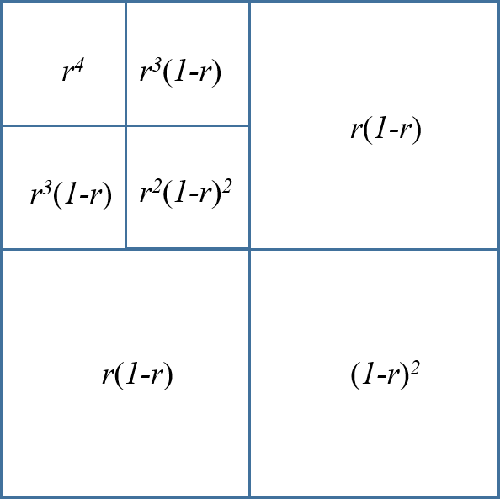}
\end{minipage}%
}%
\label{fig:figure20}
\centering
\caption{ The process of the DWT and the transformed example: (a) The process of the DWT, where downward arrows mean down-sampling. (b) The original image. (c) The result of the twp-times DWT. (d) The ratios for different patches.}
\end{figure}
\begin{equation}
I^{LL}, I^{LH}, I^{HL}, I^{HH} = {\rm DWT}(I)
\label{eq:gan}
\end{equation}
where superscripts mean the output from respective filters. {$I^{LL}$} is the down-sampling image, {$I^{HL}$} and {$I^{HL}$} are horizontal and vertical edge detection images, and {$I^{HH}$} is the corner detection image. The DWT decomposes an image into four small patches with different frequency information. Furthermore, the low-frequency information of the image (${I^{LL}}$) can be decomposed iteratively and divided into the multi-scale patches with various frequencies, as shown in Fig. 2(c). The iterative DWT can be formulated as follows:
\begin{equation}
I^{LL}_{i+1}, I_{i+1}^{LH}, I_{i+1}^{HL}, I_{i+1}^{HH} = {\rm DWT}(I_{i}^{LL})
\label{eq:gan}
\end{equation}
where subscripts mean the output of the {$i_{th}$} time DWT. It is noted that {$I_{0}^{LL}$} means the original image.

We apply DWT to improve the original SSIM loss. The SSIM for two images {$x,y$} is defined as:
\begin{equation}
  {\rm SSIM}(x,y) = \frac{(2\mu_x\mu_y + C_1)  (2 \sigma _{xy} + C_2)} 
    {(\mu_x^2 + \mu_y^2+C_1) (\sigma_x^2 + \sigma_y^2+C_2) }
  \label{eq:SSMI}
\end{equation}
where {$\mu$} and {$\sigma$} represent the means, standard deviation, and covariance of images.
A higher SSIM indicates the two images are more similar to each other, and the SSIM equals 1 for identical images. The loss function for the SSIM can be then written as follows:
\begin{equation}
  L_{{\rm SSIM}} = -{\rm SSIM}(x,y)
  \label{eq:SSMIloss}
\end{equation}

In practice, means and standard deviations are computed with a Gaussian filter. Therefore, the total function is differentiable and can be used in neural networks. The detailed derivation is analyzed in \cite{zhao2016loss}. We extend the SSIM loss function and combine the DWT. The input is not only the whole image but the various patches computed by the DWT. Varying the image resolution in a coarse-to-fine manner in loss function prevents undesired artifacts. The proposed W-SSIM loss can be described as follows:
\begin{equation}
\begin{aligned}
  &L_{{\rm W-SSIM}}(x,y) = \sum_{0}^{i}r{_i}L_{{\rm SSIM}}(x^{w}_i,y^{w}_i), \\
&w \in \left\{ LL, HL, LH, HH\right\}
\end{aligned}
\label{eq:SSMIloss}
\end{equation}
where {$r_i$} controls weights of each patch. Because hazy images suffer from low contrast, faint color, and shifted luminance, the high-frequency patches own higher ratios \cite{liu2017efficient}. We set ratios of four patches as:
\begin{equation}
  I^{LL}:I^{LH}:I^{HL}:I^{HH} = r^2:r(1-r):r(1-r):(1-r)^2
  \label{eq:SSMIloss}
\end{equation}
Initially, the SSIM loss of the whole images is computed. Then, the DWT decomposes the images, and SSIM loss of three high-frequency parts are accumulated except the low-frequency part ({$I^{LL}_{i}$}). {$I^{LL}_{i}$} can be divided by the DWT to generate various patches, and accumulate new loss. 
The Algorithm $1$ describes the weights of different patches. {$x, y$} and {$z$} are constructed as auxiliary numbers to record current weights in different scales, After the DWT is implemented, {$x, y$} and {$z$} are updated. The ratios for different patches are plotted in Fig. 2(d).
The total loss function is {$L_{2}+L_{{\rm W-SSIM}}$}. Since ratios of low-frequency patches in Algorithm $1$ are small, {$L_{2}$} is helpful to reconstruct low-frequency parts.
\label{ssec:subhead}
\begin{algorithm}
  \caption{W-SSIM Loss}
  \KwIn{Two images $I, J$, the ratio for multi-frequency $r$ and iterative times $n$}
  \KwOut{$loss = L_{{\rm W-SSIM}}(I, J)$}
  $I^{LL}_0, J^{LL}_0 = I, J$\;
  Tensor {$loss=0$}\;
  {$x=r^2, y=r(1-r), z=(1-r)^2$}\
  \For{$i=1; i \le n; i++$}
  {
    $I^{LL}_{i},I^{LH}_{i},I^{HL}_{i},I^{HH}_{i} = {\rm DWT}(I^{LL}_{i-1})$\\
    $J^{LL}_{i},J^{LH}_{i},J^{HL}_{i},J^{HH}_{i}= {\rm DWT}(J^{LL}_{i-1})$\\
    $loss+=L_{SSIM}(I^{LH}_{i}, J^{LH}_{i})\cdot y 
    +L_{{\rm SSIM}}(I^{HL}_{i},J^{HL}_{i})\cdot y+L_{{\rm SSIM}}(I^{HH}_{i}, J^{HH}_{i})\cdot z$\
    {$[ x, y, z] = x\cdot [x, y, z]$}
  }
  {$loss+=L_{{\rm SSIM}}(I^{LL}_{i},J^{LL}_{i})\cdot x$}\\
  return $loss$
\end{algorithm}
\vspace{-8mm}
\section{EXPERIMENTAL RESULTS}
\label{sec:pagestyle}
\subsection{Datasets and training details}
In this work, we adopt the RESIDE\cite{li2019benchmarking} dataset as training data. This dataset contains indoor images and outdoor images with corresponding hazy synthetic images. We mainly select 1400 outdoor images and the corresponding synthetic images as training data and select 500 images to evaluate. We set $r = 0.4$ in Algorithm 1, The 2D Haar wavelet \cite{mallat1999wavelet} is adopted, and the W-SSIM loss implements the DWT three times. During training, Adam is used \cite{kinga2015method} as the optimization algorithm with a learning rate of 0.0001, and a batch size of 32. All training images are resized to 480 {$\times$} 480, and the network is trained for 400 iterations.

\subsection{Image Dehazing Results}
We adopt PSNR and SSIM for the quantitative evaluation and compare the proposed model with several state-of-the-art dehazing methods: Color Attenuation Prior (CAP) \cite{zhu2015fast}, AOD-Net \cite{li2017aod}, Multi-band enhancement \cite{cho2018model}, and Wavelet U-net (W U-net) \cite{yang2019wavelet}. CAP and MBE are prior-based methods, and the other two methods belong to deep-learning-based methods. The networks of these two methods are the cascaded refinement network and the U-net, respectively. The comparison results are shown in Table 1. It is obvious that the PNSR and SSIM from prior-based methods are lower than other deep-learning-based methods. Furthermore, our method performs favorably against the other two deep-learning-based competitors in this dataset, which shows that the proposed Y-net can reconstruct clear and detailed images.
\begin{figure*}[ht!]
  \centering
\includegraphics[width=\linewidth]{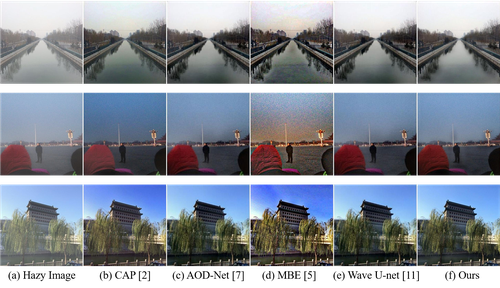}
\vspace{-1cm}
    \caption{{Dehazed results in River, People and Willow.}}
\label{fig:figure2}
\vspace{-0.5cm}
\end{figure*}

Besides quantitative analysis, a few dehazed images by all methods are depicted in Fig. 3. All images of River contain artifacts or noises on the river surface except ours. Restored images from MBE own higher contrast but are vulnerable to noise. For example, People is filled with speckle noise. We calculate fog aware density evaluation (FADE) \cite{choi2015referenceless} scores and show them in Table 2.
FADE not only provides a perceptual density score of the entire image but reports the local perceptual density of each patch. The remaining hazes, noises, and artifacts on images decrease scores. Our method produces good results on all images, which means the purposed network restores high-quality clear images.
\vspace{-3mm}
\begin{table}[ht!]
  \centering
    \caption{{Quantitative SSIM and PSNR on the synthetic RESIDE dataset.}}
    \vspace{0.6cm}
\begin{tabular}{|c|c|c|c|}
\hline
              & PNSR & SSIM  \\ \hline
CAP [2] (prior-based)& 23.02   & 0.865      \textbf{} \\ \hline
AOD-Net [7] (learning-based)      & 23.92   &  0.875     \textbf{} \\ \hline
MBE [5] (prior-based)        & 18.83   &  0.790               \\ \hline
W U-net [11] (learning-based) & 24.81     & 0.910                \\ \hline
Ours          & \textbf{26.61}     & \textbf{0.947}            \\ \hline
\end{tabular}
\end{table}
\vspace{-5mm}
\begin{table}[ht!]
  \centering
    \caption{{Quantitative FADE on restored images.}}
    \vspace{0.5cm}
\begin{tabular}{|c|c|c|c|}
\hline
              & River & People  & Willow\\ \hline
CAP [2]          & 1.41   & 0.410 &     0.496 \\ \hline
AOD-Net [7]      & 1.19   &  0.373 &    0.391 \\ \hline
MBE [5]           & 0.440   &  0.184  & 0.184            \\ \hline
W U-net [11] & 1.51     & 0.647  &     0.562         \\ \hline
Ours          & \textbf{1.77}     & \textbf{2.37} & \textbf{0.592}            \\ \hline
\end{tabular}
\end{table}

We also consider four loss functions, {$L_2$}, {$L_{{\rm SSIM}}$}, {$L_{{\rm W-SSIM}}$} and {$L_{{\rm W-SSIM}}+L_2$} to train the proposed Y-net; results are listed in Table. 3. Experimental results show that using the W-SSIM loss improves both PSNR and SSIM, even though it is an SSIM-based function. The reason is that the proposed loss function is similar to perceptual loss \cite{johnson2016perceptual}. Based on the proposed loss function, our model not only learns whole images but the different frequencies and scale features. Moreover, {$L_{{\rm W-SSIM}}+L_2$} is slightly better than original {$L_{{\rm W-SSIM}}$}.
\begin{table}[ht!]
  \centering
    \caption{{SSIM and PSNR results of all loss functions applied for the purposed network.}}
    \vspace{0.5cm}
\begin{tabular}{|lcccc|}
\hline
     & {$L_2$} & {$L_{{\rm SSIM}}$} & {$L_{{\rm W-SSIM}}$} & {$L_{{\rm W-SSIM}}+L_2$}  \\ \hline
PSNR & 26.31   & 26.27    & 26.50   & \textbf{26.61} \\ \hline
SSIM & 0.925   & 0.929   & 0.939   & \textbf{0.947} \\ \hline
\end{tabular}
\end{table}

\vspace{-10mm}
\section{CONCLUSION}
\label{sec:typestyle}
In this paper, we address the single image dehazing problem via the end-to-end Y-net that aggregates multi-scale features to reconstruct clear images. 
The experimental results demonstrate the proposed network outperforms the other existing methods, both qualitatively and quantitatively. Furthermore, the proposed W-SSIM loss is able to improve accuracy when the network architecture is unchanged.

\bibliographystyle{IEEEbib}
\bibliography{strings,refs}

\begin{thebibliography}{10}

\bibitem{mccartney1976optics}
Earl~J McCartney,
\newblock ``Optics of the atmosphere: scattering by molecules and particles,''
\newblock {\em New York, John Wiley and Sons, Inc., 1976. 421 p.}, 1976.

\bibitem{zhu2015fast}
Qingsong Zhu, Jiaming Mai, Ling Shao, et~al.,
\newblock ``A fast single image haze removal algorithm using color attenuation
  prior.,''
\newblock {\em IEEE Trans. Image Processing}, vol. 24, no. 11, pp. 3522--3533,
  2015.

\bibitem{berman2016non}
Dana Berman, Shai Avidan, et~al.,
\newblock ``Non-local image dehazing,''
\newblock in {\em Proceedings of the IEEE conference on computer vision and
  pattern recognition}, 2016, pp. 1674--1682.

\bibitem{galdran2018duality}
Adrian Galdran, Aitor Alvarez-Gila, Alessandro Bria, Javier Vazquez-Corral, and
  Marcelo Bertalm{\'\i}o,
\newblock ``On the duality between retinex and image dehazing,''
\newblock in {\em Proceedings of the IEEE Conference on Computer Vision and
  Pattern Recognition}, 2018, pp. 8212--8221.

\bibitem{cho2018model}
Younggun Cho, Jinyong Jeong, and Ayoung Kim,
\newblock ``Model-assisted multiband fusion for single image enhancement and
  applications to robot vision,''
\newblock {\em IEEE Robotics and Automation Letters}, vol. 3, no. 4, pp.
  2822--2829, 2018.

\bibitem{ren2016single}
Wenqi Ren, Si~Liu, Hua Zhang, Jinshan Pan, Xiaochun Cao, and Ming-Hsuan Yang,
\newblock ``Single image dehazing via multi-scale convolutional neural
  networks,''
\newblock in {\em European conference on computer vision}. Springer, 2016, pp.
  154--169.

\bibitem{li2017aod}
Boyi Li, Xiulian Peng, Zhangyang Wang, Jizheng Xu, and Dan Feng,
\newblock ``Aod-net: All-in-one dehazing network,''
\newblock in {\em Proceedings of the IEEE International Conference on Computer
  Vision}, 2017, vol.~1, p.~7.

\bibitem{ronneberger2015u}
Olaf Ronneberger, Philipp Fischer, and Thomas Brox,
\newblock ``U-net: Convolutional networks for biomedical image segmentation,''
\newblock in {\em International Conference on Medical image computing and
  computer-assisted intervention}. Springer, 2015, pp. 234--241.

\bibitem{yang2018auto}
C-H~Huck Yang, Fangyu Liu, Jia-Hong Huang, Meng Tian, MD~I-Hung Lin, Yi~Chieh
  Liu, Hiromasa Morikawa, Hao-Hsiang Yang, and Jesper Tegner,
\newblock ``Auto-classification of retinal diseases in the limit of sparse data
  using a two-streams machine learning model,''
\newblock in {\em Asian Conference on Computer Vision}. Springer, 2018, pp.
  323--338.

\bibitem{yang2018novel}
C-H~Huck Yang, Jia-Hong Huang, Fangyu Liu, Fang-Yi Chiu, Mengya Gao, Weifeng
  Lyu, Jesper Tegner, et~al.,
\newblock ``A novel hybrid machine learning model for auto-classification of
  retinal diseases,''
\newblock {\em arXiv preprint arXiv:1806.06423}, 2018.

\bibitem{yang2019wavelet}
Hao-Hsiang Yang and Yanwei Fu,
\newblock ``Wavelet u-net and the chromatic adaptation transform for single
  image dehazing,''
\newblock in {\em 2019 IEEE International Conference on Image Processing
  (ICIP)}. IEEE, 2019, pp. 2736--2740.

\bibitem{nah2017deep}
Seungjun Nah, Tae Hyun~Kim, and Kyoung Mu~Lee,
\newblock ``Deep multi-scale convolutional neural network for dynamic scene
  deblurring,''
\newblock in {\em Proceedings of the IEEE Conference on Computer Vision and
  Pattern Recognition}, 2017, pp. 3883--3891.

\bibitem{zhao2016loss}
Hang Zhao, Orazio Gallo, Iuri Frosio, and Jan Kautz,
\newblock ``Loss functions for image restoration with neural networks,''
\newblock {\em IEEE Transactions on Computational Imaging}, vol. 3, no. 1, pp.
  47--57, 2016.

\bibitem{wang2004image}
Zhou Wang, Alan~C Bovik, Hamid~R Sheikh, Eero~P Simoncelli, et~al.,
\newblock ``Image quality assessment: from error visibility to structural
  similarity,''
\newblock {\em IEEE transactions on image processing}, vol. 13, no. 4, pp.
  600--612, 2004.

\bibitem{liu2017efficient}
Xin Liu, He~Zhang, Yiu-ming Cheung, Xinge You, and Yuan~Yan Tang,
\newblock ``Efficient single image dehazing and denoising: An efficient
  multi-scale correlated wavelet approach,''
\newblock {\em Computer Vision and Image Understanding}, vol. 162, pp. 23--33,
  2017.

\bibitem{li2019benchmarking}
Boyi Li, Wenqi Ren, Dengpan Fu, Dacheng Tao, Dan Feng, Wenjun Zeng, and
  Zhangyang Wang,
\newblock ``Benchmarking single-image dehazing and beyond,''
\newblock {\em IEEE Transactions on Image Processing}, vol. 28, no. 1, pp.
  492--505, 2019.

\bibitem{mallat1999wavelet}
St{\'e}phane Mallat,
\newblock {\em A wavelet tour of signal processing},
\newblock Elsevier, 1999.

\bibitem{kinga2015method}
D~Kinga and J~Ba Adam,
\newblock ``A method for stochastic optimization,''
\newblock in {\em International Conference on Learning Representations (ICLR)},
  2015, vol.~5.

\bibitem{choi2015referenceless}
Lark~Kwon Choi, Jaehee You, and Alan~Conrad Bovik,
\newblock ``Referenceless prediction of perceptual fog density and perceptual
  image defogging,''
\newblock {\em IEEE Transactions on Image Processing}, vol. 24, no. 11, pp.
  3888--3901, 2015.

\bibitem{johnson2016perceptual}
Justin Johnson, Alexandre Alahi, and Li~Fei-Fei,
\newblock ``Perceptual losses for real-time style transfer and
  super-resolution,''
\newblock in {\em European conference on computer vision}. Springer, 2016, pp.
  694--711.

\end{thebibliography}

\end{document}